\documentclass[runningheads]{llncs}
\usepackage{graphicx}

\begin{document}
\title{Understanding why SLAM algorithms fail in modern indoor environments.} 
%
%
\author{Nwankwo Linus \and
Elmar Rueckert 
}
\authorrunning{L. Nwankwo \and E. Rueckert.}
%
\institute{Chair of Cyber-Physical Systems, Montanüniversität, Leoben, Austria
\url{https://cps.unileoben.ac.at/}
}
\maketitle  
\begin{abstract}
Simultaneous localization and mapping (SLAM) algorithms are essential for the autonomous navigation of mobile robots. With the increasing demand for autonomous systems, it is crucial to evaluate and compare the performance of these algorithms in real-world environments. In this paper, we provide an evaluation strategy and real-world datasets to test and evaluate SLAM algorithms in complex and challenging indoor environments. Further, we analysed state-of-the-art (SOTA) SLAM algorithms based on various metrics such as absolute trajectory error, scale drift, and map accuracy and consistency. Our results demonstrate that SOTA SLAM algorithms often fail in challenging environments, with dynamic objects, transparent and reflecting surfaces. We also found that successful loop closures had a significant impact on the algorithm's performance. These findings highlight the need for further research to improve the robustness of the algorithms in real-world scenarios.
\end{abstract}


\section{Introduction}
\begin{figure}[ht]
 \centering
  \subfigure[Hospital]{\includegraphics[scale=0.22]{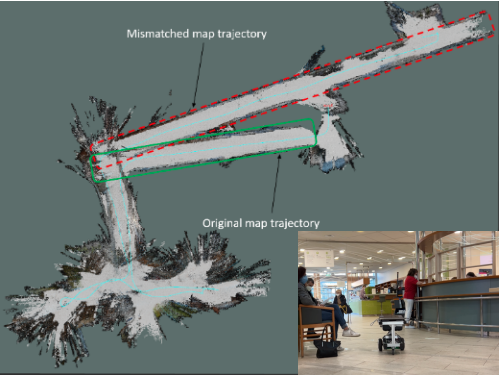}}
  \subfigure[Study Center]{\includegraphics[scale=0.12]{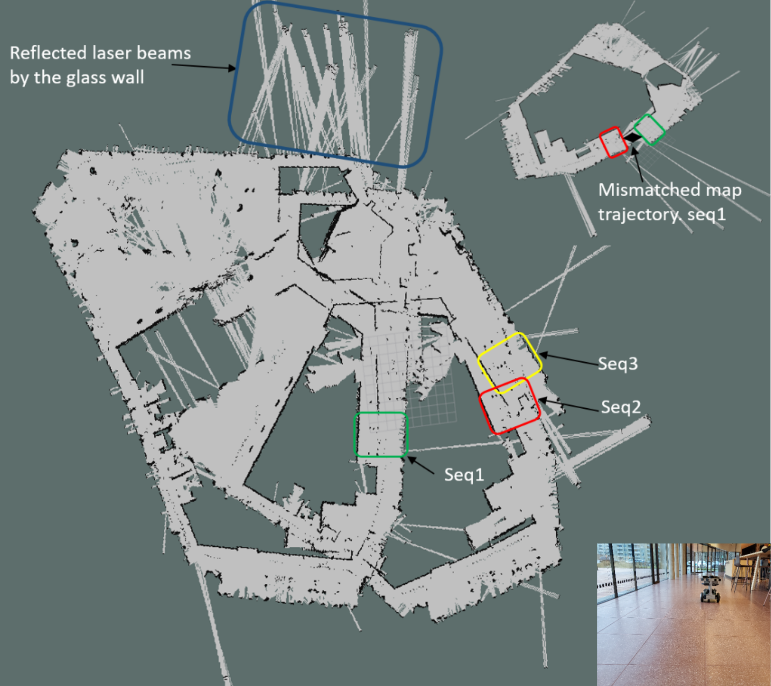}}
  \subfigure[CPS Lab.]{\includegraphics[scale=0.27]{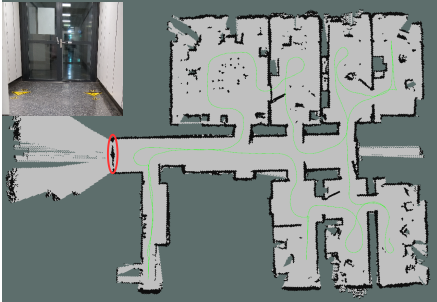}}
  \subfigure[Long Passage]{\includegraphics[scale=0.490, angle =90]{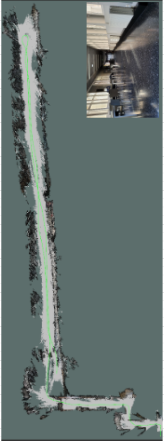}}
  \subfigure[Controlled Env.]{\includegraphics[scale=0.190, angle =0]{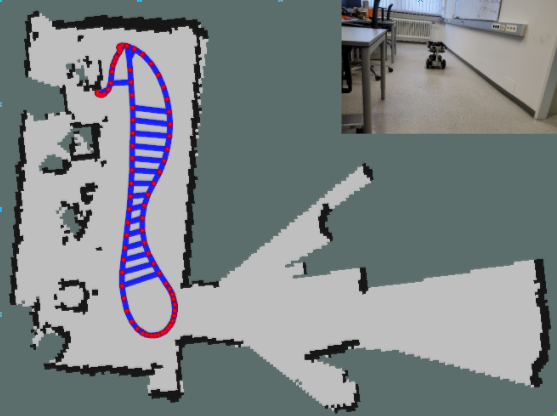}}
 \caption{Illustration of the environments. In the areas marked in red in (a) and (b), the algorithm could not track the robot's trajectory. The algorithm failed to create a consistent metric map as a result of the accumulation of odometry error and insufficient geometric features for the pose estimation due to the reflectivity of glass walls.}
    \label{fig:o2senvirons}
\end{figure}

SLAM algorithms have been widely researched in the past decade and have shown significant progress in simulated and laboratory environments \cite{slamFuturePresent}. However, the real-world deployment of these algorithms is still a challenging task due to factors such as poor odometry, dynamic objects, transparent surfaces e.g., glass walls, etc.
In real-world environments, SLAM algorithms face several challenges such as illumination variations, geometric-free regions, limited sensor range, etc. These challenges often cause the algorithm to fail in providing accurate and consistent estimates of the robot's position and the environment's structure.
For example, in Figure \ref{fig:o2senvirons}a-b, the algorithms failed to create a consistent metric map of the environments due to the glass walls reflectivity, different lightening conditions, unpredictable obstacles, no geometric features at the open entrance, and problems attributed to odometry error. The glass walls (Figure \ref{fig:o2senvirons}b) reflect the laser beams which therefore return inaccurate measurements.

These challenges limits the applicability of the SLAM algorithms in various fields such as autonomous navigation, robotic mapping, and augmented reality. Therefore, it is important to understand the sources of failure and to develop new methods to overcome them.

\section{Related Work}
Most SLAM algorithms are based on vision (with monocular, stereo, or RGBD cameras) or lidar (2D or 3D) \cite{10.48550/arxiv.2207.00254}. Previous studies have shown that traditional SLAM algorithms, such as Kalman Filter-based \cite{inproceedingsRAAD}, particle filter-based \cite{Stachniss2016}, and graph-based \cite{graph-based-slam} approaches struggles in challenging environments due to their reliance on accurate sensor data and assumptions about the environment. Recent research has focused on developing more robust SLAM algorithms that can better handle such environments. For example, Xiangwei Dang  et al.\cite{sensorFusion} proposed incorporating sensor fusion techniques, where multiple sensors are used to improve the robustness and accuracy of the algorithm.

Additionally, a study \cite{li} have proposed using visual-based SLAM algorithms, which rely on visual information from cameras, and can be more robust in dynamic environments. However, these algorithms still have limitations in certain scenarios such as low lighting conditions or when the camera is occluded.

\section{Method}
To ascertain the root cause of failures of SLAM algorithms, we set up a series of tests in controlled and challenging real-world scenarios that included factors such as reflecting surfaces, different lightening conditions, high levels of noise, and dynamic or unpredictable obstacles. In this section, we describe the hardware platforms, the features of the SLAM algorithms used for this work. Thereafter, we provide details of the test environments, datasets, evaluation metrics, and our experimental setup. 

\subsection{Hardware Platforms}
We used our developed low cost mobile robot\cite{linusetal} with two standard high-resolution RGB-D cameras, a $360^\circ$ horizontal field of view (FOV) lidar sensor, and a 9-axis inertial measurement unit (IMU) to record novel challenging datasets used to evaluate the robustness and accuracy of the selected SLAM algorithms. 
Other hardware platforms used for this test include a Lenovo ThinkBook Intel Core i7, Intel iRIS Graphics running on Ubuntu 20.04, ROS Noetic distribution with 8GB RAM. These were used for on-the-site (OTS) data-set acquisition. Additionally, since the recorded datasets are processed offline on each of the considered algorithms, we used a ground station PC with Nvidia Geforce RTX 3060 Ti GPU, 8GB GDDR6 memory running Ubuntu 20.04 and ROS Noetic distribution.

\subsection{SLAM Algorithms}
We evaluated SOTA open-source SLAM algorithms actively supported and  maintained by  the  long-term  end-of-life (EOL)  distribution  of the robot operating system (ROS) \cite{ros}. Due to the large availability of open-source SLAM algorithms, we limited our selection to the most popular and well-supported (a) three filter and graph-based SLAM approaches namely: Hector-SLAM \cite{hector-slam}, Gmapping \cite{gmap}, and Karto-SLAM \cite{karto} and (b) two visual-based techniques namely: real-time appearance-based mapping (RTAB-Map) \cite{rtap-map} and ORB-SLAM2 \cite{orb-slam}. 

These algorithms were run on the same datasets recorded in each of the test environments, and their performances are assessed based on the evaluation metrics in Subsection \ref{sec2E}.

\subsection{Environment and Data Collection}
Our datasets were recorded in five complex indoor environments of different geometric and feature characteristics depicting realistic operations scenarios for autonomous mobile robots. 
Figure \ref{fig:o2senvirons} shows the exemplary layout of the environments. The CPS Lab (Figure \ref{fig:o2senvirons}c) depicts a typical office layout with multiple occlusions, several pieces of furniture,  and different lighting conditions (i.e., natural daylight and artificial lights). The Long Passage (Figure \ref{fig:o2senvirons}d) is a typical passageway of approximately $6$ x $120\;m$ with several tables and natural day lighting conditions. Figure \ref{fig:o2senvirons}b is the student study centre at the University of Leoben with large reflecting surfaces and many moving objects (people) which depicts a typical static and dynamic environment. The hospital environment (Figure \ref{fig:o2senvirons}a) includes both static, dynamic characteristics (i.e., people moving around, some sitting, and several pieces of furniture) as well as multiple reflecting glasses. The last dataset was collected in a room with limited obstruction and dynamics which we considered as controlled environment (Figure \ref{fig:o2senvirons}e). 

We captured the dataset by teleoperating the mobile robot around the environments leveraging the ROS-Mobile framework developed by \cite{nils}. We recorded specific topics of interest, e.g., for lidar-based SLAM algorithms, the robot's transformation, odometry, and scan data are required. For vision-based technique, the RGB-D dataset contains the colour and depth images of an Intel Realsense D435i camera along the ground-truth trajectory of the sensor. The data was recorded at a full frame rate of 30 frames per second (fps), RGB and depth resolution of $1920$ x $1080$ and $1280$ x $720$ respectively. The ground-truth trajectory was obtained from both Intel Realsense $T265$ tracking camera and a high-accuracy external $9$ degree-of-freedom (DOF) motion tracking IMU device mounted on the robot base.

\subsection{Metrics}\label{sec2E}
We assessed the performance and the accuracy of the algorithms based on the metrics described in this subsection. To ensure a detailed performance analysis, we performed both quantitative and qualitative evaluations.
For the quantitative assessment, we defined various metrics relating to the robot's trajectory and translation (Equations \ref{eqnATE}, \ref{eqnRPE_trans}, \ref{eqnRPE_rot}, and \ref{eqnSD}).  Furthermore, for the qualitative evaluation, we performed a visual inspection of the generated map and compared it to the ground truth map.

\textbf{Absolute trajectory error (ATE):}
We utilised this metric to measure the overall alignment between the estimated trajectory and the ground truth trajectory.
The ATE error directly reflects the accuracy of the estimation algorithm and the consistency of the global trajectory. 
More formally, given a  sequence of poses from the estimated trajectory ${s}_{i} \triangleq [{s}^{trans}_{i}, {s}^{rot}_{i}]^{T} \in SE(3)$ and the ground truth trajectory $g_{i} \triangleq [{g}^{trans}_{i}, {g}^{rot}_{i}]^{T} \in SE(3)$ at the $i-th$ timestamp, we computed the ATE as shown in Equation \ref{eqnATE}. $SE(3)$ denotes the homogeneous transformation of the robot in 3D space \cite{mueller2019modern}. The superscripts $trans$ and $rot$ denotes the translational and rotational components of the transformation matrix respectively.
\begin{equation}\label{eqnATE}
    ATE_{trans} \triangleq \sqrt{\frac{1}{n} \sum_{i=1}^{n} {\norm{g^{trans}_{i} - {s}^{trans}_{i}}}^{2}} 
\end{equation}
Where $||.||$ is the $L(2)$ norm of the position difference of the Euclidean distance between ground truth poses $g_{i}$ and the estimated poses ${s}_{i}$. The variable $n$ is the total number of poses in the trajectory sequence.

\textbf{Relative pose error (RPE):}
We choose this error metric as complementary to ATE since it does not account for the accumulated error of the robot's trajectory. However, it enabled us to independently estimate the drift in each of the SLAM algorithms. We computed RPE as the average of the Euclidean distance between the $g_{i}$ and the $s_{i}$, normalized by the $g_{i}$ to ensure that the RPE is expressed as a ratio, rather than an absolute value i.e.,
\begin{equation}\label{eqnRPE_trans}
    RPE_{trans} \triangleq \sqrt{\frac{1}{m} \sum_{i=1}^{m} {\norm{\frac{g^{trans}_{i} - {s}^{trans}_{i}}{g^{trans}_{i}}}^{2}}}\\ 
\end{equation}
where $m$ is the number of relative poses in the trajectory. 
We used Equation \ref{eqnRPE_trans} and \ref{eqnRPE_rot} to compute the difference in position and orientation between the consecutive estimated and ground truth poses respectively.
\begin{equation}\label{eqnRPE_rot}
    RPE_{rot} \triangleq \frac{1}{n} \sum_{i=1}^{n} (cos^{-1}(g^{rot}_{i} - {s}^{rot}_{i}))
\end{equation}

\textbf{Scale Drift (SD):}
We defined this metric to measure the deviation of the algorithm's estimated scale of the environment to the ground truth scale over time. It enabled us to measure the consistency of the estimated trajectory with respect to the true scale. We computed the scale drift by comparing the cumulative distance of the estimated pose with the cumulative distance of the ground truth pose i.e.,
\begin{equation}\label{eqnSD}
  SD = \frac{1}{n-1}\sum_{i=1}^{n-1}\frac{\norm {(s^{trans}_{i+1} - s^{trans}_{i})}}{\norm{(g^{trans}_{i+1} - g^{trans}_{i})}} 
\end{equation}
From \ref{eqnSD}, if the SLAM algorithm is performing optimally (i.e., with no deviation), then SD = 1, and when it deviates from the actual values, SD values less than or greater than 1 indicate underestimation or overestimation of the distances. 

\textbf{Map Quality:}
We used this metric to measure the quality of the map generated by each SLAM algorithm. The map quality was evaluated by comparing the generated map to a ground truth map,  and by assessing the consistency, correctness and completeness of the map through visual inspection.

\section{Results}
In this section, we present the statistical results of the performance of the different algorithms on our datasets. Each algorithm was evaluated on the same recorded dataset with its default settings to ensure a fair comparison.

\subsection{Lidar-based and Visual-based SLAM Techniques}
Table \ref{Tab: results} presents the evaluation results of each of the SLAM algorithms considered in this work. The performances are benchmarked on our recorded datasets from the five indoor environments. We grouped the results into lidar and visual based. For each group and environment, we represent the best results for the metric in \textbf{bold} font.
\begin{table*}[h]
\caption{Statistical results of the algorithms performance with the different recorded datasets. $\mu \rightarrow$ mean, $\sigma \rightarrow$ standard deviation, $t/a \rightarrow $ translation /angular RMSE.}
  \centering
\begin{tabular}{l|c|c|c|c||c|c|}
\hline
\multicolumn{2}{c|}{\multirow{2}{*}{Metrics}} & \multicolumn{3}{c||}{Lidar-Based} & \multicolumn{2}{c|}{Visual-Based}\\
\cline{3-7}
\multicolumn{1}{l}{} & & Hector & Gmapping & Karto & ORBSLAM2 & RTAB\\
\hline
\multirow{4}{*}{\begin{turn}{90}Hospital\end{turn}}
                               & ATE($\mu/\sigma$) & \textbf{0.12/0.09} & 0.28/0.26  & 0.29/0.26 &   8.51/8.76 & \textbf{8.49/8.76}\\
                               & RPE($\mu/\sigma$) & \textbf{0.10/0.35} & 0.12/0.78  &  0.13/0.79 &  4.80/29.36  &  \textbf{0.53/0.45}\\
                               & SD($\mu/\sigma$) & 1.50/0.38 &  \textbf{1.02/0.05}  & 1.03/ 0.06 &  \textbf{{1.59}/3.50}  &  \textbf{{1.59}/3.50}\\
                               & RMSE (t/a) & \textbf{0.16/0.38} & 0.39/0.13   & 0.39/0.14 &  \textbf{12.22/1.48}  & 12.23/1.49\\
\hline
\multirow{4}{*}{\begin{turn}{90}St. Center\end{turn}} 
                               & ATE($\mu/\sigma$) & \textbf{0.15/0.01} & 0.19/0.01  & 0.17/0.03 &  3.59/2.37 &  \textbf{0.25/0.20}\\
                               & RPE($\mu/\sigma$) & \textbf{0.06/0.64} & 0.08/0.82  & 0.07/0.73 &  0.32/0.91  &  \textbf{0.08/0.84}\\
                               & SD($\mu/\sigma$) & 1.01/0.02 & 1.04/0.02   & \textbf{0.99/0.02} &  0.67/0.16  &  \textbf{0.81/0.05 }\\
                               & RMSE (t/a) & \textbf{0.15/0.13} & 0.19/0.12   & 0.17/0.12 &  4.30/0.51  & \textbf{0.32/0.08}\\
\hline
\multirow{4}{*}{\begin{turn}{90}CPS Lab.\end{turn}} 
                               & ATE($\mu/\sigma$) & 1.73/1.14 & 1.77/1.36  & \textbf{1.54/1.18} &  \textbf{0.75/0.60} &  0.87/0.53\\
                               & RPE($\mu/\sigma$) & 0.25/0.59 & \textbf{0.21/0.56}  & 0.22/0.59 &  \textbf{0.10/0.16}  &  0.32/2.68\\
                               & SD ($\mu/\sigma$) & 1.05/0.16 & 1.03/0.07  &  \textbf{0.97/0.13} &  \textbf{1.03/0.03}  &  1.13/0.99\\
                               & RMSE (t/a) & 2.08/0.34 & 2.23/0.36   & \textbf{1.94/0.71} &  \textbf{0.96/0.60}  & 1.02/0.53\\
\hline                             
\multirow{4}{*}{\begin{turn}{90}Long Pass.\end{turn}} 
                              & ATE($\mu/\sigma$) & \textbf{0.06/0.04} & 1.32/0.77  & 1.57/0.92 &  0.13/0.05 & \textbf{0.06/0.06}\\
                               & RPE($\mu/\sigma$) & \textbf{0.01/0.03} & 0.08/0.23  & 0.09/0.23 &  0.48/4.79  &  \textbf{0.01/0.06}\\
                               & SD ($\mu/\sigma$) & 0.96/0.01 & \textbf{0.98/0.02}  & \textbf{0.98/0.02} &  1.18/2.14  &  \textbf{1.17/2.14}\\
                               & RMSE (t/a) & \textbf{0.07/0.12} & 1.53/0.39   & 1.82/0.43 &  0.13/0.11  & \textbf{0.08/0.11}\\
\hline   
\multirow{4}{*}{\begin{turn}{90}Ctrld. Env.\end{turn}} 
                              & ATE($\mu/\sigma$) & 0.29/0.24 & \textbf{0.01/0.00}  & 0.02/0.00 &  \textbf{0.31/0.25} &  0.32/0.30\\
                               & RPE($\mu/\sigma$) & 0.46/0.62 & \textbf{0.03/0.08} & 0.10/0.25 & 0.43/0.67  &  \textbf{0.35/0.62}\\
                               & SD ($\mu/\sigma$) & 1.75/0.49 & \textbf{1.00/0.00}   & \textbf{1.00/0.00} &  0.70/0.21  &  \textbf{0.71/0.20}\\
                               & RMSE (t/a) & 0.38/0.47 & \textbf{0.01/0.00}  & 0.02/0.00 &   \textbf{0.40/0.31}   & 0.44/0.22\\
\hline
\end{tabular}
\label{Tab: results}
\end{table*}
From Table \ref{Tab: results}, all the algorithms had difficulty dealing with the complexity of the environments, resulting in large ATE, RPE, and RMSE (as evident in the hospital environment for visual-based SLAM technique, and CPS Lab for lidar-based approach).

Comparing the individual lidar-based algorithms, Hector-SLAM outperforms all other methods in the Long Passage environment with low mean and low standard deviation. This is attributed to its unreliance on the error-prone odometry data to perform accurate scan matching. Also, the cluttered  static obstacles at environment enabled it to perform accurate scan matching and generate consistent occupancy grid map of the environment.

Gmapping performed sufficiently at the controlled environment. We attributed its poor performance in other environments to its reliance on perfect odometry to maintain a set of hypotheses about the robot's pose. Odometry is inherently uncertain and prone to error due to the complexity of the operating environments and factors such as wheel slippage, sensor noise, and drift over time. The performance in the controlled environment could also be attributed to the fact that the robot covered less distance and thus accrued fewer odometry errors.

Karto-SLAM had near optimal SD values in the majority of the environments (fourth column) due to its graph-based SLAM approach, which adapts well to changes in the environment.
However, the high mean and standard deviation indicates that the algorithm still struggles to estimate the robot's pose accurately, leading to larger errors.

\begin{figure}[ht]
 \centering
 \vspace{-.3cm}
 \subfigure[Controlled Env.]{\includegraphics[scale=0.225]{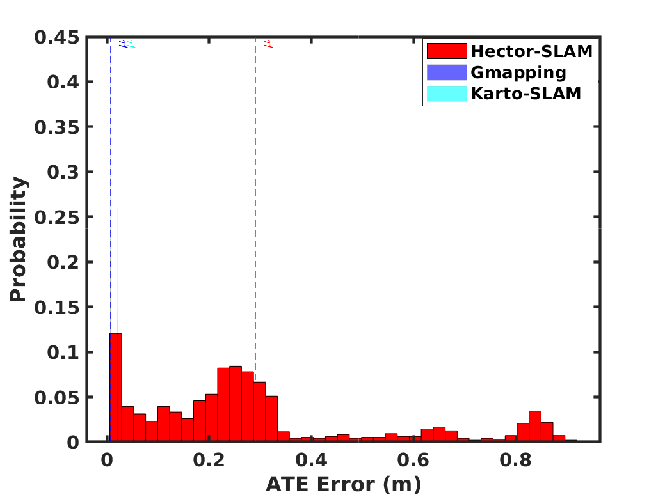}}
 \subfigure[CPS Lab.]{\includegraphics[scale=0.225]{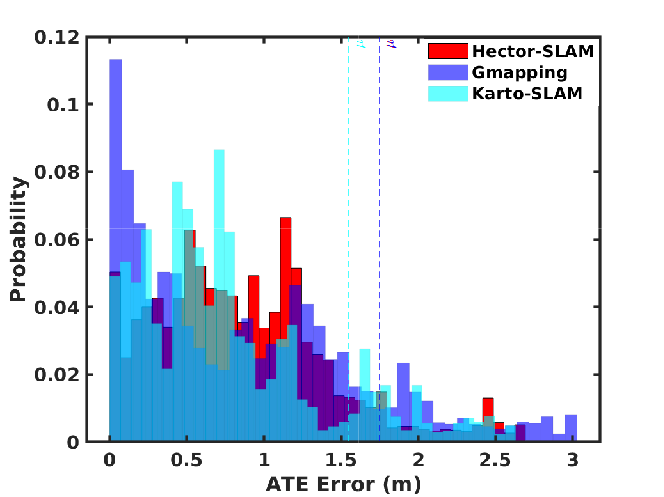}}
 \subfigure[Hospital]{\includegraphics[scale=0.225]{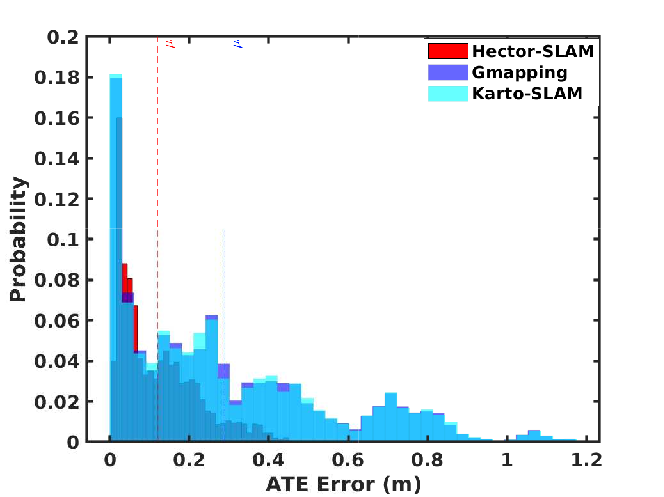}}
 \subfigure[Study Center]{\includegraphics[scale=0.225]{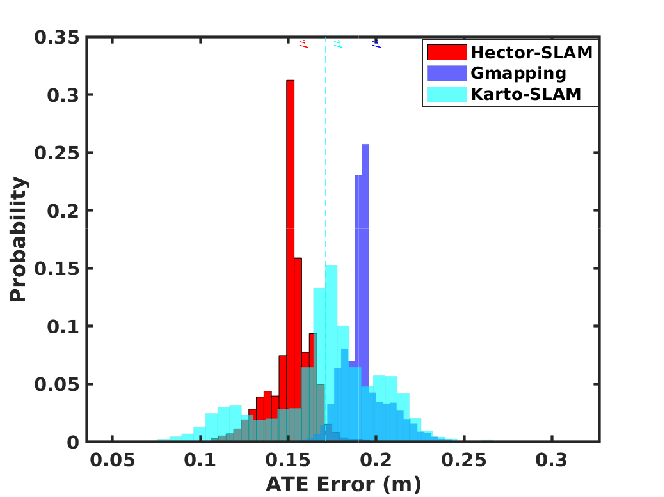}}
 \subfigure[Long Passage]{\includegraphics[scale=0.225]{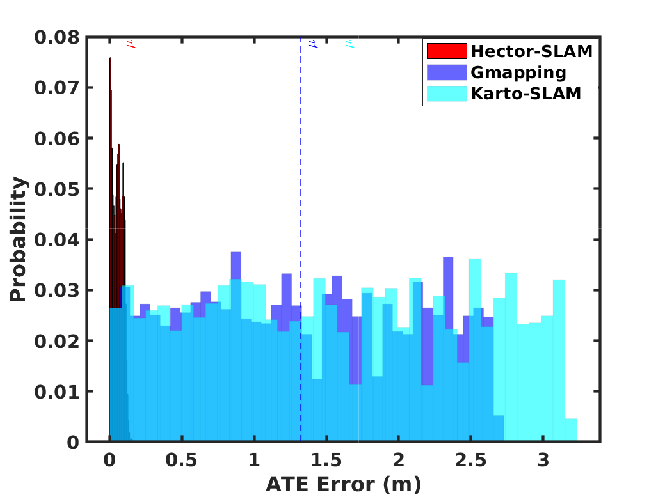}}
 \subfigure[Visual Long Passage]{\includegraphics[scale=0.22]{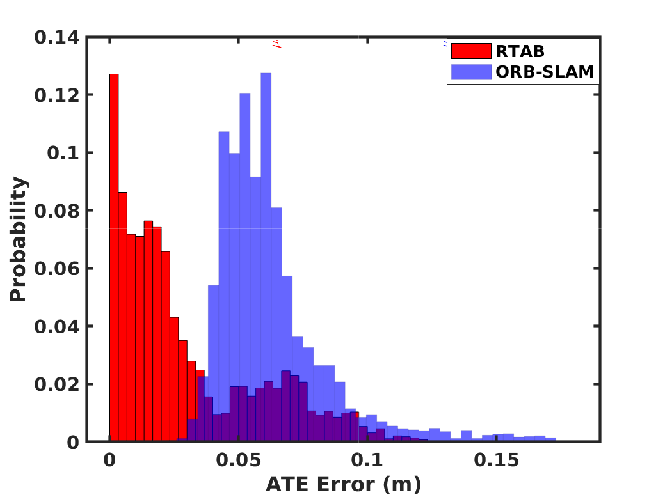}}
 \caption{ATE result for all the environments. In each plot, the dashed vertical lines represent the mean ATE for each algorithm. }\label{Fig: ATE}
\end{figure}

Figure \ref{Fig: ATE} shows the distribution of the ATE for the algorithms. In the hospital environment (Figure \ref{Fig: ATE}c), the ATE for Gmapping algorithm  has a probability of 0.18 of being between $0.00 m$ and $1.18 m$. This means that for $18\%$ of the time, the error in the estimated trajectory is within that range. Furthermore, the ATE and the mean translational and angular root mean square error $RMSE (t/a)$ for the CPS Lab are high for all SLAM algorithms due to frequent turning, loop closures, unpredictable obstacles, and the limited geometric features in the environment. The mean ATE of $1.77 m$ e.g., indicates that if at any moment, the robot has to return to a specific location in the environment, there is at least $1.77 m$ of error to reach it, without considering the accumulation of such error. 

\section{Conclusion}
In this work, we provided an evaluation strategy and real-world datasets to evaluate SOTA SLAM algorithms in controlled and in  challenging indoor environments. In the controlled environment, the majority of the algorithms performed adequately, with relatively low ATE, RPE, and RMSE values e.g., ATE mean of the lidar based Gmapping algorithm was $0.01$m and of the visual SLAM approach ORBSLAM2 was $0.31$m. However, in challenging environments, the ATE, RPE, RMSE and SD values were higher e.g., the ATE mean of Gmapping was $1.77$m and of RTAB $0.87$m in the CPS lab environment. The complete evaluation results of five algorithms in five environments is shown in Table~\ref{Tab: results}.

These results demonstrate the limitations of the algorithms in complex and dynamic environments, where the robot has to navigate through obstacles (see Figure~\ref{fig:o2senvirons} a), changing lighting conditions, and dealing with dynamic objects. The results were influenced by factors such as the reflecting surfaces (see Figure~\ref{fig:o2senvirons} b), sensor noise, odometry errors, and the large-scale of the environment. In summary, we have shown that the SLAM algorithms face significant challenges in challenging environments and that further research and development are needed to improve their robustness in real-world scenarios.

To improve the performance, we are exploring the incorporation of  floor plans as prior to complement the SLAM map in current ongoing work. 

%
%
 \bibliographystyle{splncs04}
 \bibliography{reference}
\end{document}